\documentclass[final]{cvpr}

\usepackage{times}
\usepackage{epsfig}
\usepackage{graphicx}
\usepackage{amsmath}
\usepackage{amssymb}

\usepackage{setspace}
\usepackage{booktabs}
\usepackage{array}
\usepackage{multirow}

\usepackage[pagebackref=true,breaklinks=true,colorlinks,bookmarks=false]{hyperref}

\begin{document}

\title{2D+3D Facial Expression Recognition via Discriminative Dynamic Range Enhancement and Multi-Scale Learning}

\author{
Yang Jiao $^{\dagger, \ddagger}$, Yi Niu$^{\dagger}$, Trac D. Tran $^{\ddagger}$, Guangming Shi $^{\dagger}$\\
$^{\dagger}$ Xidian University, $^{\ddagger}$ Johns Hopkins University\\
}
\maketitle

\begin{abstract}
    In 2D+3D facial expression recognition (FER), existing methods generate multi-view geometry maps to enhance the depth feature representation. However, this may introduce false estimations due to local plane fitting from incomplete point clouds.
    In this paper, we propose a novel Map Generation technique from the viewpoint of information theory, to boost the slight 3D expression differences from strong personality variations. First, we examine the HDR depth data to extract the discriminative dynamic range $r_{dis}$, and maximize the entropy of $r_{dis}$ to a global optimum. Then, to prevent the large deformation caused by over-enhancement, we introduce a depth distortion constraint and reduce the complexity from $O(KN^2)$ to $O(KN\tau)$. Furthermore, the constrained optimization is modeled as a $K$-edges maximum weight path problem in a directed acyclic graph, and we solve it efficiently via dynamic programming. Finally, we also design an efficient Facial Attention structure to automatically locate subtle discriminative facial parts for multi-scale learning, and train it with a proposed loss function $\mathcal{L}_{FA}$ without any facial landmarks. Experimental results on different datasets show that the proposed method is effective and outperforms the state-of-the-art 2D+3D FER methods in both FER accuracy and the output entropy of the generated maps.
\end{abstract}

\section{Introduction}
\label{sec:Intro}
Facial expression -- the most primary and straightforward nonverbal way to convey the emotion of a human being -- plays an essential role not only in human-behavior analysis \cite{2007_Kapoor, RealEyes_web} but also in human-machine interaction \cite{2012_Tan, 2014_Blom, 2014_DeVault}.
In recent years, 3D and 2D+3D based Facial Expression Recognition (FER) have attracted much attention to overcome non-ideal lighting conditions and large pose variations from RGB data, since the depth and texture are somewhat complementary for expression representation.

However, even with the aid of 3D information, multi-modality FER is still challenging since facial features are dominated by strong personality (intra-class) differences, but not by slight expression (inter-class) variations. For example, muscle movements from Angry to Sadness only cause minor inter-class differences because of the invariance of facial structure. However, diversity of independent subjects (e.g., gender, age, race) introduces large intra-class variations to expression data. Therefore, enhancing weak expression features from strong personality variations is the key for successfully recognition. To this end, \textit{Map Generation} has been designed as a critical and indispensable stage in the 2D+3D FER pipeline, where raw 3D scans are comprehensively represented by a series of attribute maps.

\begin{figure}[htbp]
\begin{center}
    \includegraphics[width=0.85\linewidth]{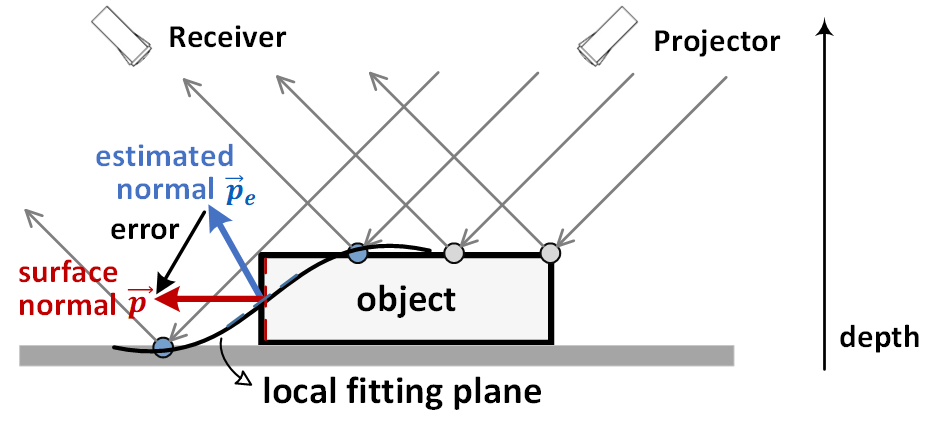}
\end{center}
    \caption{False normals are introduced by local plane fitting.}
\label{fig:falseNormal}
\vspace{-2mm}
\end{figure}

\textbf{Problem.} In the current literature \cite{2015FG_Yang, 2015CVIU_Li, 2016TMM_Zhen, 2017_YangCNN, 2017TMM_DFCNN, 2018FG_Wei, 2019FG_DACNN,   2019SP_FERLrTc}, geometrical information is widely explored to describe raw 3D scans for map generation. For instance, Yang \etal \cite{2017_YangCNN} generates curvature maps (CuM) from 3D face model to enhance the facial landmarks, and Zhen \etal \cite{2016TMM_Zhen} extracts normal maps (NoM) and shape index maps (SIM) values to describe the geometry deformation of facial regions. Recently, diverse geometry maps (GeM) are combined together to further promote the feature learning in deep neural networks as adopted in \cite{2017TMM_DFCNN, 2018ACMMM_Chen, 2019FG_DACNN}.
These methods provide elegant directions to enhance 3D expression features from multi-geometry views. However, considering that in real FER applications, depth data are only acquired once from the view of frontal face. This one-shot setup can only sample an incomplete 3D data of the scene, missing the point clouds that might be essential for geometry estimation. In other words, the generated multi-view geometry maps in existing methods are NOT projected by the real 3D objects, but estimated by local plane fitting \cite{LocalPlaneFitting, LocalCubicFitting}. Unfortunately, fitting a surface plane with incomplete data may bring false geometrical attributes to the generated map as illustrated in Fig. \ref{fig:falseNormal}, and decreases recognition performances. In the figure, the estimated normal $\mathop{p_{e}}\limits ^{\rightarrow}$ (blue arrow) from local fitting plane introduces the error (black arrow) to the real surface normal $\mathop{p}\limits ^{\rightarrow}$ (red arrow).

\textbf{Motivation.} Taking a different approach from current works, we are inspired by the fact that depth scans are sampled by high dynamic range (HDR) point clouds, while inter-class differences are only represented by a limited dynamic range, indicating a low signal-to-noise ratio (SNR) of raw 3D expression scans. Hence, enhancing the low dynamic range discriminative features of HDR data could provide an effective route for better expression representation. Also, considering that entropy is a natural tool to characterize the ability of exhibiting the richness of image content, maximizing the entropy of the expression data is a reasonable approach to enhance weak discriminative features.
Following this observation, we propose a novel coarse-to-fine Depth Distortion Constraint based Globally Optimal Entropy Maximization (GEMax) technique to pursue the maximal amount of information from the discriminative dynamic range for FER with multi-scale feature representation.

\textbf{Method.} The proposed GEMax contains three stages: S1) Range Selection, S2) Depth Distortion Constraint based Globally Optimal Entropy Maximization and S3) Data Collection. In stage S1, we examine the full dynamic range to preserve critical features in $D_{dis}$ with discriminative range $r_{dis}$, and eliminate the expression-irrelative data at a coarse level. Then in stage S2, entropy of $D_{dis}$ is maximized to represent the optimal information at a fine level. To overcome the face deformation caused by over-enhancement in unconstrained tone mapping, we introduce a depth distortion constraint $\tau$, and significantly reduce the complexity from $O(KN^2)$ to $O(KN\tau)$ as well. We model the constrained optimization problem as finding a $K$-edges maximum weight path in a directed acyclic graph (DAG), and obtain the optimal solution with efficient dynamic programming. Finally, in stage S3, generated maps with different $\tau$ are collected together to jointly represent the facial expression and train the network. Here, we also design an effective Facial Attention (FA) structure with FA loss $\mathcal{L}_{FA}$ to automatically locate discriminative facial parts for multi-scale feature learning. FA can be trained without extra facial landmark annotations.

\textbf{Original contributions} are summarized as follows.
\vspace{-1mm}
\begin{itemize}
\item We demonstrate the effectiveness of enhancing raw point clouds for FER from the viewpoint of information theory. To the best of our knowledge, this is the first work using non-geometrical strategy for map generation in deep learning based 2D+3D FER methods.

\item We propose a novel map generation technique GEMax for 2D+3D FER by maximizing the globally optimal entropy of the discriminative dynamic range with a new depth-distortion constraint.

\item We design an efficient Facial Attention (FA) structure to automatically locate local facial parts for multi-scale learning. FA can be trained with facial attention loss $\mathcal{L}_{FA}$ without any facial landmarks supervision.
\vspace{-1mm}
\end{itemize}

Extensive experimental results show that our method significantly improves the entropy of the generated 3D maps from geometrical based methods: from 7.47 \cite{2019FG_DACNN, 2018FG_Wei} to 7.94. The proposed technique also outperforms state-of-the-art approaches in different 2D+3D FER benchmarks: BU-3DFE ($89.72\%$ vs. $88.35\%\cite{2019FG_DACNN}$) and Bosphorus ($83.63\%$ vs. $82.50\%$\cite{2018FG_Wei}). Finally, ablation studies reveal the effectiveness of each step in the proposed method.

\section{Related Works}
\label{sec:RelatedWorks}

FER methods can be divided into 2D FER, 3D FER and 2D+3D FER from the aspect of data modality. In this paper, we focus on 3D/2D+3D FER techniques, in which depth features are enhanced and represented by \textit{Map Generation}.

\subsection{Facial Expression Recognition (FER)}
\textbf{3D FER.} The earlier literature adopts a variety of handcrafted features for 3D FER tasks, such as Scale-Invariant Feature Transform (SIFT) in \cite{2011_Berretti, 2010_Berretti}, and Histogram Of Gradient (HOG) in \cite{2015CVIU_Li, 2011ACIVS_Li}. Recently, deep learning based methods learn 3D features by designing end-to-end Convolutional Neural Networks (CNN) with the following pipeline: 1) map generation; 2) feature extraction; 3) feature fusion; and 4) expression recognition.
Li \etal \cite{2015_Li_3DFER} first converts depth data to five geometric maps, then adopts a pre-trained VGG-19 \cite{VGG} followed by linear SVMs, achieving accuracy level of $84.87\%$ on BU-3DFE \cite{BU3DFE}. Yang \etal \cite{2017_YangCNN} generates facial attribute masks from landmarks to supervise the training of CNN model and achieves $75.90\%$ on BU-4DFE \cite{BU4DFE}. More recently, Zhu \etal \cite{2019FG_DACNN} proposes a Discriminative Attention-based CNN (DA-CNN) method with an accuracy level of $87.69\%$, in which salient regions are estimated by CNN to enhance the local facial features. Different from \cite{2019FG_DACNN}, Chen \etal \cite{2018ACMMM_Chen} enhances facial 3D surface by proposing a novel human vision inspired pooling structure via Fast and Light Manifold CNN (FLM-CNN) to recognize expressions directly from point-cloud data. However, merely launching FER from 3D data is still insufficient due to the lack of rich texture information.

\textbf{2D+3D FER.} By utilizing both texture and depth scans for comprehensively inter-class difference representation, multi-modality FER techniques achieve higher gains than single-modality FER. Fu \etal \cite{2019SP_FERLrTc} proposes a low-rank tensor completion (FERLrTC) to capture multi-linear low-rank structure and achieves $82.89\%$ on BU-3DFE.
Li \etal \cite{2017TMM_DFCNN} proposes a Deep Fusion CNN (DF-CNN) to combine feature learning and fusion via a unified end-to-end framework. It is the first work that introduces deep CNN and feature-level fusion to 2D+3D FER. Then, DA-CNN from Zhu \etal \cite{2019FG_DACNN} is also designed to implicitly learn the salient multi-modality facial maps and it achieves $88.35\%$ on BU-3DFE. Instead of learning full image features, Jan \etal \cite{2018FG_FP} estimates 4 local facial parts from both 2D and 3D data with facial landmarks for feature fusion, achieving an impressive accuracy of $88.54\%$ on BU-3DFE.

Different from the aforementioned methods, we solve 2D+3D FER task by enhancing the slight inter-class variations for 3D point clouds, and jointly learn the local-global multi-scale features via the Facial Attention structure.

\subsection{Map Generation}
Map Generation is critical in 3D and 2D+3D FER by representing raw depth scans with a series of 2D attribute maps in order to: 1) enhance discriminative depth features; and 2) generate uniform grids for network input.

Uniform quantization is the most straightforward way to generate low dynamic range images as adopted in \cite{2008_HaoTang, 2009_Gong, 2017ICCVws_Oyedotun, 2018FG_FP, 2019MMSP_MultiView}. It is simple and fast, but may cause potential information loss for small expression variations. Hence, Vieriu \etal \cite{2015FG_Cylindrical} projects 3D data onto a cylindrical 2D plane for wide-rage rotational invariance. Normal patterns and curvature components are used to describe the first order and second order local structure in \cite{2012_Li} and \cite{2013_Lemaire, 2016TMM_Zhen}. Li \etal \cite{2011ACIVS_Li, 2015CVIU_Li} also adopts shape index maps for surface representation. In recent years, multi-view strategy is widely used to fully exploit the geometry properties via local plane/cubic fitting methods \cite{LocalPlaneFitting, LocalCubicFitting} for deep learning based FER, such as \cite{2017TMM_DFCNN, 2017_YangCNN, 2018FG_FP, 2018ACMMM_Chen, 2018FG_Wei, 2019FG_DACNN}. For example, curvature maps, range images and facial masks are concatenated to form 3-channel data to augment expressions in \cite{2017_YangCNN}. Along this direction, Li \etal \cite{2015_Li_3DFER} introduces a novel map generation by using 5 different attribute maps, namely a geometry map, 3 normal maps and a normalized curvature map for joint feature learning. Similar techniques are also adopted in \cite{2018FG_Wei, 2019FG_DACNN}.

Instead of exploring multi-view geometrical attributes, we construct a more effective map generation process from the viewpoint of information theory. Specifically, we propose to maximize the entropy of the discriminative dynamic range while restricting the large depth distortion.

\section{Method}
\label{sec:3}
In the proposed method, raw HDR depth scans $D$ are comprehensively represented by a series of LDR depth images via three stages: S1) Range Selection, S2) Depth Distortion Constraint based Globally Optimal Entropy Maximization and S3) Data Collection. After stages S1 and S2, diverse maps under different depth constraints are collected in the S3 stage. Once accomplished, multi-scale feature learning via Facial Attention structure is employed in the final recognition stage.

\subsection{Range Selection (S1)}
\label{sec:3.1}
\begin{figure}[htbp]
\begin{center}
    \includegraphics[width=0.85\linewidth]{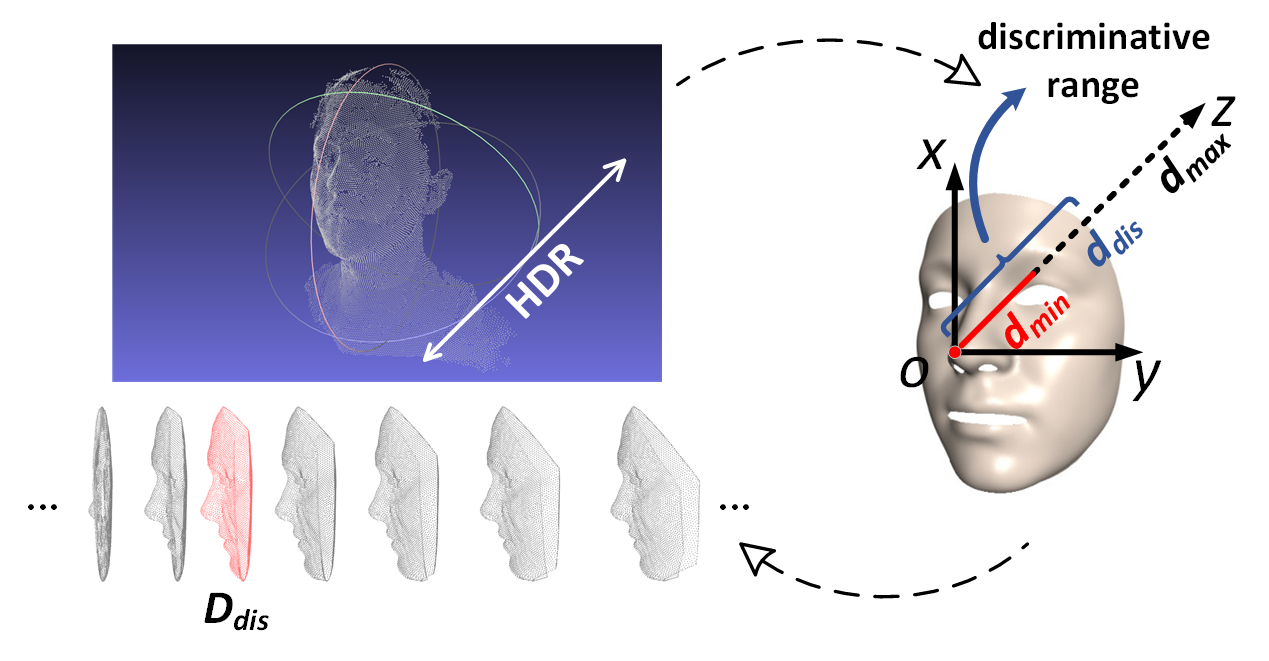}
\end{center}
   \caption{Illustration of Range Selection.}
\label{fig:depthSelection}
\vspace{-2mm}
\end{figure}

We design Range Selection to extract the discriminative range $r_{dis}$ from raw 3D data to preserve critical features while eliminating expression-irrelevant structures. Considering $r_{dis} = d_{dis}/r_z$, computing $r_{dis}$ equates to finding the discriminative depth $d_{dis}$ under fixed-depth resolution $r_z$. An intuitive direction is to decompose the raw point clouds $D$ into a series of depth block $\{D_i\}$ with different depth, then evaluate each $D_i$ independently to examine its contribution to the FER task. We search the raw data from depth $d_{min}$ cm to $d_{max}$ cm, starting from the nose as illustrated in Fig. \ref{fig:depthSelection}, and generate overlapped depth blocks with incremental $\Delta d$. Here, $d_{min}$ is the lower bound to keep the basic face structure and $d_{max}$ is the maximum sampled depth.

By evaluating $D_i$, we discover that point clouds data from different dynamic ranges contribute significantly unevenly to FER task. The recognition gap between the best depth block $D_{dis}$ (with $d_i=6$ cm) and the worst ($d_i=14$ cm) is higher than $2.5\%$ which is a huge gain in the current 2D+3D FER task. This observation reflects that the most discriminative expression features are located on limited shallow depth with limited dynamic range $r_{dis}$, and motivates us to further enhance the discriminative features for $r_{dis}$ on a finer level.

\subsection{Depth Distortion Constraint based Globally Optimal Entropy Maximization (S2)}
\label{sec:3.2}

\begin{figure*}[htbp]
\begin{center}
    \includegraphics[width=0.85\linewidth]{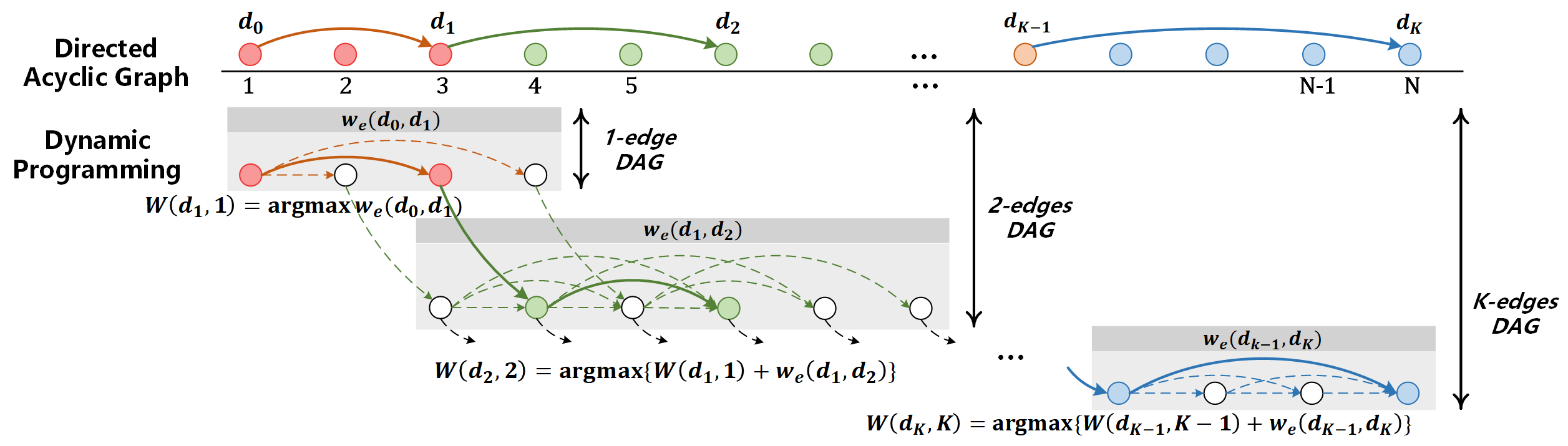}
\end{center}
   \caption{Dynamic programming process for solving $K$-edges maximum weight path optimization in DAG with depth distortion constraint.}
\label{fig:DynamicProgramming}
\vspace{-4mm}
\end{figure*}

The S2 stage takes the HDR depth block $D_{dis}$ from the S1 stage as input, and produces LDR depth images $I_d$ with enhanced features. Let $N$, $K$, and $K < N$ denote the number of the input and output depth levels, then mapping function from $D_{dis}$ to $I_d$ can be defined by an ordered integer vector
\begin{equation}
\textbf{F} = \{d_0, d_1, ..., d_k, ..., d_K\},
\label{eq:mapping_function}
\end{equation}
where $d_0 = 0$ and $d_K = N-1$. In $\textbf{F}$, any input depth data from range $[d_k, d_{k+1})$ will be mapped to the low dynamic intensity level $k$. With this setting, the objective function can be written as maximizing the entropy of $I_d$ from $D_{dis}$ with the mapping function $\textbf{F}$ as
\vspace{-1mm}
\begin{equation}
\sum\nolimits_{k}-\textit{P}[d_k, d_{k+1})log\textit{P}[d_k, d_{k+1}),
\vspace{-2mm}
\end{equation}
where $\textit{P}[d_k, d_{k+1}) = \sum_{i=d_k}^{d_{k+1}}p_i$ is the probability of intensity $k$ in LDR $I_d$. This can be calculated by accumulating $p_i$ from $i=d_k$ to $d_{k+1}$ in HDR depth data $D_{dis}$.

To solve this optimization problem, Histogram Equalization (HE) based methods with a global optimal solution seem to yield reasonable solutions. However, HE may over-enhance the depth image $I_d$ by merging too many point clouds from $D_{dis}$ with low probabilities. This will cause severe geometrical deformation and disrupt the face structure. Therefore, we introduce a depth distortion constraint condition rather entropy alone for map generation.

\textbf{Depth Distortion Constraint.}
We define the depth distortion for each compressed bins $k$ as
\vspace{-1mm}
\begin{equation}
 \varepsilon_k = (d_{k+1} - d_k) / N,
\vspace{-2mm}
\end{equation}
where $(d_{k+1} - d_k)$ denotes the number of bins in $D_{dis}$ that would be mapped to $k$ in $I_d$, and $N$ is the dynamic range of $D_{dis}$ for normalization. Considering that any bins of $D_{dis}$ between range $[d_k, d_{k+1})$ would be mapped into one single depth value in the output depth image $I_d$, $\varepsilon_k$ is a natural metric to evaluate the distortion loss of the mapping process. In this work, we constrain the maximal value of $\varepsilon_k$ by enforcing $\varepsilon_k < \tau$ for any $k$ in entropy maximization. Hence, the following additional constraint should be considered in the objective function:
\vspace{-1mm}
\begin{equation}
1 \leq d_{k+1} - d_k \leq \tau, \text{for any } k.
\label{eq:depth_constraint}
\vspace{-1mm}
\end{equation}
Finally, the objective of the S2 stage can be formulated as the following constrained optimization problem:
\vspace{-1mm}
\begin{equation}
\begin{aligned}
&\textbf{F} = \underset{\textbf{F}}{\mathrm{argmax}} \sum_{k=0}^{K-1}-\textit{P}[d_k, d_{k+1})log\textit{P}[d_k, d_{k+1})\\
&s.t.\quad
\begin{cases}
0 \leq k \leq K-1, \\
0 \leq d_k \leq N-1. \\
1 \leq d_{k+1} - d_k \leq \tau, \text{for any } k.
\end{cases}
\end{aligned}
\label{eq:object_function}
\end{equation}

It should be noticed that the value of depth distortion constraint $\tau$ should be in the range of $[N/K, N]$. If $\tau=N/K$, the map generation process becomes uniform quantization from $N$ to $K$, known as range image. And if $\tau=N$, the new constraint will contribute nothing to the solution.

\subsection{Graph Model and Dynamic Programming}
\label{sec:3.3}
To solve the constrained optimization in (\ref{eq:object_function}), an intuitive way is to greedy search the cumulative function such as HE, but this restricts the maximum searching length for every step to $\tau$. However, the greedy strategy is NP hard (combinatorial) and might fall into locally optimal solutions. In pursuit of the globally optimal solution, we re-examine the constrained problem via graph theory, and propose a dynamic programming based solution.

\begin{figure}[htbp]
\begin{center}
    \includegraphics[width=1.0\linewidth]{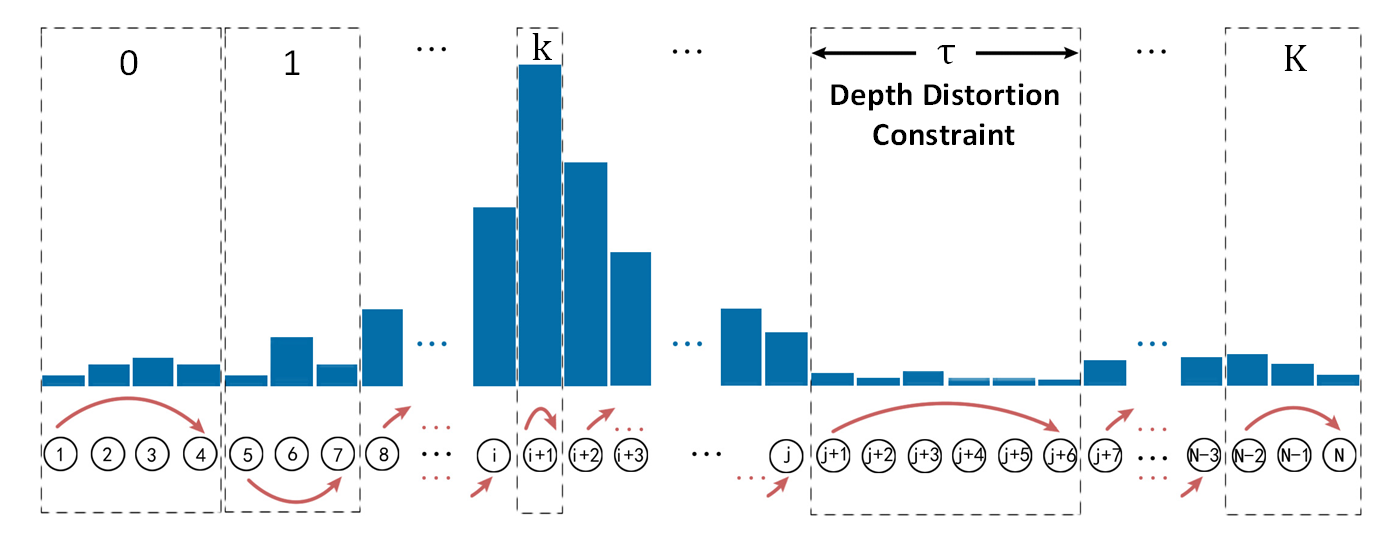}
\end{center}
    \caption{Illustration of graph theoretical based representation of the proposed map generation.}
\label{fig:GraphModel}
\vspace{-2mm}
\end{figure}

\textbf{Graph model. } First, let us define the graph model, where the input histogram of $D_{dis}$ can be modeled by a directed acyclic graph (DAG) denoted by $G(V, E)$. The nodes of the DAG vertex set $V = \{1, 2, ..., N\}$ represent the input depth levels $d_k$, and the path between any two nodes $[i, j)$ with $0 \leq i < j \leq N$ indicates the edge $e(i, j) \in E$. The weight of the edge $e(i, j)$ denoted by $w_e(i, j)$ represents the cost of mapping a depth range $[i, j)$ from $D_{dis}$ to a single pixel in $I_d$ as $w_e(i, j) = -\textit{P}[i, j)log\textit{P}[i, j)$. With this set-up, the mapping function $\textbf{F}$ in (\ref{eq:mapping_function}) corresponds to a $K$-edges path in $G(V, E)$ linking node $0$ to $N$, and the objective function in (\ref{eq:object_function}) is equivalent to seeking the optimal $K$-edges path in $G(V, E)$ with maximum weights $w_e(i, j)$ under the depth distortion constraint from (\ref{eq:depth_constraint}). Considering that the depth distortion is defined by the maximum number of the combined bins in the mapping function $\textbf{F}$, the constraint (\ref{eq:depth_constraint}) can be satisfied by not allowing any edge from node $i$ to $j$ if $j - i > \tau$ in the construction of $G(V, E)$  as illustrated in Fig. \ref{fig:GraphModel}. In this fashion, the constrained optimization problem (\ref{eq:object_function}) can be rewritten as the following unconstrained equivalence:
\vspace{-2mm}
\begin{equation}
W(N, K) = \underset{\textbf{F}}{\mathrm{argmax}} \sum_{k=0}^{K-1}w_e(d_k, d_{k+1})
\label{eq:unconstraint_version}
\vspace{-2mm}
\end{equation}
where $W(N, K)$ is any path in $G(V, E)$ starting from the node $0$ and ending at node $N$ with $K$ edges determined by $\textbf{F}$.

However, solving the unconstrained optimization problem in (\ref{eq:unconstraint_version}) is still NP-hard since we have to emulate all possible paths in $G(V, E)$. Therefore, we investigate an efficient dynamic programming solution for (\ref{eq:unconstraint_version}), which can be performed in polynomial time.

\textbf{Dynamic programming solution.} Considering that if $d_0 \to d_K$ is the $K$-edges maximum weight path from vertex $0$ to $N$, then the subpath $d_0 \to d_{K-1}$ must be the $(K-1)$-edges maximum weight path from $0$ to $d_{K-1}$. With this divide-and-conquer strategy, the original optimization problem of $W(N, K)$ in (\ref{eq:unconstraint_version}) can be represented by a series of sub-problems as shown in the following recursive form:
\vspace{-2mm}
\begin{equation}
\begin{split}
W(N, K)
&= \underset{\textbf{F}}{\mathrm{argmax}} \sum_{k=0}^{K-2} w_e(d_k, d_{k+1}) + w_e(d_{K-1}, d_K) \\
&= \underset{d_{K-1}}{\mathrm{argmax}} (W(d_{K-1}, K-1) + w_e(d_{K-1}, d_K)) \\
&= ... \\
&= \underset{d_1}{\mathrm{argmax}} (W(d_1, 1) + \sum_{k=1}^{K-1}w_e(d_k, d_{k+1})).
\end{split}
\label{eq:subproblems}
\vspace{-2mm}
\end{equation}

To iteratively solve (\ref{eq:subproblems}), we first solve the 1-edge maximum weight path problem $W(x, 1)$ subject to $0 < x < \tau$, which is exactly the weight of edge $e(0, x)$, as the beginning step of dynamic programming. Then, in the next iteration, we focus on the 2-edge maximum weight path task $W(x, 2)$ and obtain the optimal solution by traversing all the possible intermediate vertices $v$ via
\vspace{-2mm}
\begin{equation}
W(x, 2) = \underset{v}{\mathrm{argmax}} (W(v, 1) + w_e(v, x)).
\label{eq:2edges_problem}
\vspace{-2mm}
\end{equation}
Similarly, 3-to-$K$-edges maximum weight path problems $W(x, 3)$ to $W(x, K)$ can be solved consequently by repeating (\ref{eq:2edges_problem}) multiple times until we exhaust the maximum weight path $W(N, K)$ as Fig. \ref{fig:DynamicProgramming} illustrates.

\textbf{Complexity analysis.} The dynamic programming process can be regarded as filling a table with $K \times N$ cells. The computation of any cell $W(X, Y)$ needs to traverse all possible intermediate vertices from 1 to $X-1$. Hence, the overall complexity of solving $W(N, K)$ is $O(KN^2)$. But in the proposed method, the maximum length of edge in graph $G(V, E)$ is bounded by the depth constraint $\tau$. Therefore, only $\tau$ vertices from $X-\tau$ to $X-1$ need to be traversed, reducing the complexity from $O(KN^2)$ to $O(KN\tau)$.

\subsection{Facial Attention for Multi-Scale Learning}
\label{sec:3.4}
\begin{figure}[htbp]
\begin{center}
    \includegraphics[width=0.95\linewidth]{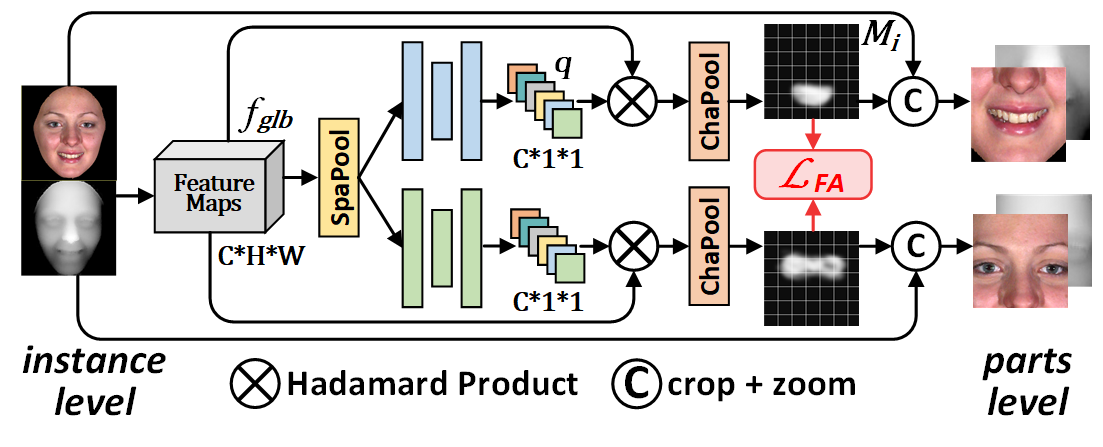}
\end{center}
    \caption{Structure of Facial Attention.}
\label{fig:FAmodule}
\vspace{-2mm}
\end{figure}

To further extract subtle local features, we design an additional Facial Attention module to automatically locate the discriminative facial parts in multi-scale learning without requiring any extra landmarks. This FA structure is depicted in Fig. \ref{fig:FAmodule}.

In the FA structure, global features $f_{glb}$ with size $C*H*W$ from texture image $I_t$ or generated 3D maps $I_d$ are extracted by a series of convolutional layers. Then, a spatial average pooling (SpaPool) layer followed by a fully-connected based auto-encoder structure is constructed to project the feature map $f_{glb}$ to a query vector $q$ with size $C*1*1$. In the query vector, we force the value from $q$ to indicate the importance of the corresponding channel in $f_{glb}$. Hence, we can repeat $q$ cross spacial domain, and implement the Hadamard product to re-weight the original global feature $f_{glb}$, and then average the feature intensities of the result for the same receptive filed via a channel average pooling (ChaPool) layer to obtain the facial attention map $M=\{M_i\}$, $i=1, ..., N_M$.

To train the FA network, auxiliary classifiers are constructed after the ChaPool layer as the only supervision step without any extra parts annotations. Unfortunately, only class-wise labels are not enough due to the two channels in FA may fall into the same local minima, resulting in the non-distinctive facial parts between $M_i$. Hence, we design the following Facial Attention loss $\mathcal{L}_{FA}$ to constrain: 1) the diversity of the discriminative parts; and 2) the concentration of each $M_i$ as follows.
\begin{equation}
\begin{split}
\mathcal{L}_{FA} = \sum_{i=1}^{N_M} \sum_{(x, y)}
& \alpha M_i(x, y)\cdot (\max_{j \neq i}  M_j(x, y) - mrg) +\\
& \beta M_i(x, y) \cdot (||x-t_x||^2 + ||y - t_y||^2).
\end{split}
\label{eq:loss_att}
\end{equation}
In this loss function, the first term evaluates the diversity between any two attention map $M_i$ whereas the second term measures the local distribution around peak magnitude $(t_x, t_y)$ for each $M_i$. The two terms are balanced by parameters $\alpha$ and $\beta$, and $mrg$ makes $\mathcal{L}_{FA}$ robust to noise. Next, the total loss $\mathcal{L}$ is then the sum of cross entropy loss and FA loss and can be expressed as $\mathcal{L} = \lambda_{CE} \mathcal{L}_{CE} + \lambda_{LA} \mathcal{L}_{LA}$.

The FA module can be embedded in any existing 2D+3D FER CNNs. In our method, we construct the framework by integrating the proposed GEMax and FA module for multi-scale learning. More specifically, enhanced depth maps are firstly generated by GEMax, then the FA module is used to automatically predict the local facial parts. Both full images and generated parts are jointly learned by VGG-16, and finally all features are fused and then fed into the classifier for the final prediction. More structure details are illustrated in the Supplementary Materials.

\section{Experiments}
\label{sec:4}

\subsection{Dataset and Experiment Protocal}
\label{sec:4.1}

\textbf{BU-3DFE} \cite{BU3DFE} (Binghamton University 3D Facial Expression) benchmark includes 100 subjects for 2D+3D static FER. Each subject has six basic expressions (i.e. Angry, Disgust, Fear, Happy, Sad and Surprise) with four mild-to-strong intensity levels and one neural expression, resulting a total $100\times(4\times6+1)=2500$ texture-depth data pairs. We follow the common protocol used in \cite{2015CVIU_Li, 2017TMM_DFCNN, 2017WAPR_OTMFA, 2018FG_FP, 2018ACMMM_Chen, 2019FG_DACNN} and use the highest 2 intensity levels for training and testing. In training, a partition of 60 subjects is randomly selected out of 100 people, repeating 100 rounds independently. In each round, 54 subjects (with $54\times2\times6=648$ pairs) are used for training and validation, and the remaining 6 (72 pairs) are for testing. 10-fold cross validation is utilized in each round for more stable performances.

\textbf{Bosphorus} \cite{BOS} is a relatively small but more challenging face database for multi-modality FER tasks. It contains a total 4,666 2D texture and 3D depth image pairs from 105 subjects under adverse conditions, such as diverse poses and occlusions. But only 65 people contain all six prototype expressions. Followed by a common protocol as that of \cite{2017TMM_DFCNN, 2018FG_Wei, 2019MMSP_MultiView, 2019SP_FERLrTc}, 60 samples are randomly selected from 65 subjects. Similarly to BU-3DFE, we repeat each round 100 times for comprehensively stable accuracy, and design a 10-fold cross validation for each round. Specifically, in each round, 90\% data ($60 \times 0.9 \times 6=324$ image pairs) are used for training and validation, and the remaining 10\% samples are reserved for testing.

\subsection{Implementation Details}
\label{sec:4.2}

\textbf{Network structure.}
To be consistent with previous methods \cite{2017TMM_DFCNN, 2019FG_DACNN, 2018FG_FP}, we employ VGG-16 \cite{VGG} as the backbone. For FA module, the number of attention map $N_M$ is set to 2, which means two different facial parts will be generated for each sample. The auto-encoder structure is designed with 512-64-512 FC layers followed by ReLU activations. Global feature map $f_{glb}$ is extracted from the last convolutional layer of VGG-16 with size [512, 7, 7], and the size of attention map $M_i$ is $7 \times 7$. Discriminative facial parts from the FA module are cropped from the input texture and depth data to size $96 \times 96$, and then upsampled to the fixed size $224 \times 224$. After the FA module, all the generated discriminative facial parts as well as full images are fed into VGG-16 for joint feature extraction, and then be fused to a single feature map with size $[6 \times 512, 7, 7]$ for final expression classification.

\textbf{Training.}
All input 2D-3D pairs are resized to $224 \times 224$. We train the network from scratch with batch size 32, and adopt the Adam optimizer with momentum [0.9, 0.999] and weight decay 0.001 for 30 epochs optimization. Learning rate is initialized to 0.1, and decreased by a factor of 10 every 10 epochs. For the loss function, we set $\lambda_{CE}=1.0$, $\lambda_{LA}=1.0$, and $\alpha=1e3$ and $\beta=1e2$ in (\ref{eq:loss_att}). For GEMax, the depth range $d_i$ is selected as 6 cm for the stage of S2, and the depth distortion constraint $\tau$ is set to $\tau={20, 30, 40, 50, 60}$ for training and $\tau=20$ for testing.

\begin{figure*}[htbp]
\begin{center}
    \includegraphics[width=1.0\linewidth]{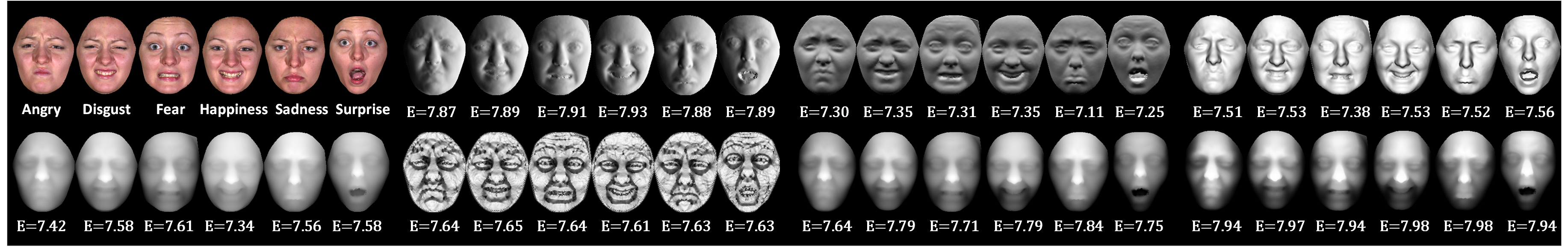}
\end{center}
   \caption{Map generation comparisons. First row from left to right: RGB images, Normal Maps on x-axis, y-axis and z-axis. Second row: Range Images, Curvature Maps, GEMax (S1), GEMax (S2). Our method achieves the highest entropy (E) with complete facial structure.}
\label{fig:CompareMaps}
\vspace{-2mm}
\end{figure*}

\subsection{Accuracy Comparison}
\label{sec:4.3}

3D and 2D+3D FER methods with and without extra facial landmark annotations are compared in Tab. \ref{tab:FERresults_BU3DFE} and Tab. \ref{tab:FERresults_BOS} for BU3DFE and Bosphorus, respectively. In these tables, 'Map' means the map generation process, and the entropy of the generated maps is evaluated in the column labeled 'Entropy'. 'Feat.' and 'Acc.' stand for model feature and accuracy. The accuracy levels in the brackets indicate that the model is trained with additional annotations, such as facial landmarks from Facial Action units (FACs) system. To have a comprehensive comparison, we also provide an additional variant model denoted by (-GeM), which is trained using geometry maps.

\begin{table}[htbp]
\small
\begin{spacing}{1.13}
\begin{center}
\setlength{\tabcolsep}{1.0mm}{
    \caption{FER comparisons on BU-3DFE dataset. }
\vspace{-4mm}
\label{tab:FERresults_BU3DFE}
\begin{tabular}{cccccc}
    \bottomrule[1.5pt]
    Method & Domain & Map & Feat. & Entropy & Acc.(\%)\\
    \hline
    Gong \etal \cite{2009_Gong}          & 3D & RI  & BFSC & 7.21 & (76.22) \\
    Berretti \etal \cite{2010_Berretti}  & 3D & RI  & SIFT & 7.21 & (77.54) \\
    Zeng \etal \cite{2013_Zeng}          & 3D & CFI & SC   &  -   & 70.93 \\
    Li \etal \cite{2012_Li}              & 3D & NoM & LNP  & 7.57 & 80.14 \\
    Li \etal \cite{2011ACIVS_Li}         & 3D & CuM & HOG  & 7.63 & (82.01) \\
    Lemaire \etal \cite{2013_Lemaire}    & 3D & CuM & HOG  & 7.63 & 76.61 \\
    Zhen \etal \cite{2016TMM_Zhen}       & 3D & CuM & HOG  & 7.63 & 83.20 \\
    Yang \etal \cite{2015FG_Yang}        & 3D & GeM & SIFT & 7.47 & 84.80 \\
    Chen \etal \cite{2018ACMMM_Chen}        & 3D & GeM & CNN  & 7.47 & (86.67) \\
    Zhu \etal \cite{2019FG_DACNN}           & 3D & GeM & CNN  & 7.47 & 87.69 \\

    \hline

    Fu \etal \cite{2019SP_FERLrTc}  & 2D+3D & GeM & LBP & 7.47 & 82.89 \\
    Li \etal \cite{2015CVIU_Li}     & 2D+3D & CuM & HOG & 7.63 & (86.32) \\
    Jan \etal \cite{2018FG_FP}      & 2D+3D & RI  & LBP & 7.21 & (86.89) \\
    Li \etal \cite{2017TMM_DFCNN}     & 2D+3D & GeM & CNN & 7.47 & 86.86 \\
    Wei \etal \cite{2018FG_Wei}   & 2D+3D & GeM & CNN & 7.47 & 88.03 \\
    Zhu \etal \cite{2019FG_DACNN}      & 2D+3D & GeM & CNN & 7.47 & 88.35 \\
    Jan \etal \cite{2018FG_FP}      & 2D+3D & RI  & CNN & 7.21 & (88.54) \\
    \hline
    ours(-GeM)                 & 2D+3D & GeM   & CNN & 7.47 & \bf 89.11\\
    ours                 & 2D+3D & GEMax & CNN & \bf 7.94 & \bf 89.72\\
    \toprule[1.5pt]

\end{tabular}}
\end{center}
\end{spacing}
\vspace{-8mm}
\end{table}

For BU-3DFE dataset as shown in Tab. \ref{tab:FERresults_BU3DFE}, our method achieves the best $89.72\%$ accuracy for 2D+3D FER, surpassing the SOTA CNN based method \cite{2019FG_DACNN} $1.37\%$. Though \cite{2018FG_FP} achieves slightly better performance than \cite{2019FG_DACNN}, it uses 49 facial landmarks for training as the additional supervision for 2D-3D feature locating. Almost all of the CNN methods \cite{2017TMM_DFCNN, 2018ACMMM_Chen, 2018FG_Wei, 2019FG_DACNN} use geometry maps (GeM) for map generation, but the entropy of GeM (7.47) is lower than the proposed GEMax (7.94), indicating a lower amount of represented information. For traditional methods, diverse maps are used to describe the feature of depth in both 3D and 2D+3D tasks, including Range Image (RI), Curvature Maps (CuM), Normal Maps (NoM) and Conformal Factor Images (CFI), but the performances are lower than CNN models and our method.

Similar results from Bosphorus benchmark can be found in Tab. \ref{tab:FERresults_BOS}, where our method again consistently achieves SOTA accuracies $83.05\%$ and $83.63\%$ with geometry maps and GEMax, surpassing all CNN and hand-crafted based competitors.

\begin{table}[htbp]
\small
\begin{spacing}{1.15}
\begin{center}
\setlength{\tabcolsep}{1.0mm}{
    \caption{FER comparisons on Bosphorus dataset.}
\label{tab:FERresults_BOS}
\begin{tabular}{cccccc}
    \toprule[1.5pt]
    Method & Domain & Map & Feat. & Entropy & Acc.(\%)\\
    \hline

    Li \etal \cite{2012_Li}           & 3D & NoM & LNP  & 7.52 & 75.83 \\
    Yang \etal \cite{2015FG_Yang}     & 3D & NoM & SIFT & 7.52 & 77.50 \\

    \hline

    Fu \etal \cite{2019SP_FERLrTc}    & 2D+3D & GeM & LBP & 7.48 & 75.93 \\
    Li \etal \cite{2015CVIU_Li}        & 2D+3D & SIM & HOG & 7.65 & (79.72) \\
    Tian \etal \cite{2019ISSC_DFFCNN}  & 2D+3D & NoM & CNN & 7.52 & 79.17 \\
    Li \etal \cite{2017TMM_DFCNN}         & 2D+3D & GeM & CNN & 7.48 & 80.00 \\
    Li \etal \cite{2017TMM_DFCNN}         & 2D+3D & GeM & CNN & 7.48 & 80.28 \\
    Wei \etal \cite{2018FG_Wei}       & 2D+3D & GeM & CNN & 7.48 & 82.22 \\
    Vo \etal \cite{2019MMSP_MultiView} & 2D+3D & RI & CNN & 7.13 & 82.40 \\
    Wei \etal \cite{2018FG_Wei}       & 2D+3D & GeM & CNN & 7.48 & 82.50 \\

    \hline
    ours(-GeM)     & 2D+3D & GeM & CNN & 7.48 & \bf 83.05\\
    ours           & 2D+3D & GEMax & CNN & \bf 7.92 & \bf 83.63\\
    \bottomrule[1.5pt]
\end{tabular}}
\end{center}
\end{spacing}
\vspace{-6mm}
\end{table}

\subsection{Entropy Evaluation}
\label{sec:4.4}
GEMax significantly improves the entropy of depth data from Range Image (RI) $7.21 \rightarrow 7.94$ and $7.13 \rightarrow 7.92$ for BU-3DFE and Bosphorus in Tab. \ref{tab:FERresults_BU3DFE} and Tab. \ref{tab:FERresults_BOS} (column labeled 'Entropy'). Widely used Geometrical Maps (GeM) have the average entropy of 7.47 by combining RI, NoM and CuM. Expression-level entropy for map generation are illustrated in Fig. \ref{fig:CompareMaps}, in which our method achieves the highest expression entropy with complete face information.

\subsection{Running Time}
\label{sec:4.5}
Since there is no parallel operation, GEMax can be easily launched on CPU machine without GPUs. Here, we use a single Intel Core i5-8250U CPU. Notice that map generation can be performed before network training and testing, and GEMax will not affect the inference time of CNN.

\begin{table}[htbp]
\small
\begin{center}
    \caption{Time-Entropy evaluation of GEMax.}
\label{tab:Ab_time}
\begin{tabular}{p{1.3cm}<{\centering} p{0.45cm}<{\centering} p{0.45cm}<{\centering} p{0.45cm}<{\centering}  p{0.45cm}<{\centering}  p{0.45cm}<{\centering}  p{0.45cm}<{\centering}  p{0.8cm}<{\centering} }

\toprule[1.5pt]
$\tau$ & 16 & 20  & 30  & 40  & 50  & 60  & $N$  \\
\hline
time (s)  & 0.21  & 0.22  & 0.23  & 0.25  & 0.29  & 0.30  & 33.9  \\
FPS       & 4.67  & 4.44  & 4.24  & 3.89  & 3.44  & 3.29  & 0.029 \\
Entropy   & 7.606 & 7.811 & 7.931 & 7.964 & 7.980 & 7.987 & 7.993 \\
\bottomrule[1.5pt]

\end{tabular}
\end{center}
\vspace{-6mm}
\end{table}

Tab. \ref{tab:Ab_time} evaluates the running time on BU-3DFE dataset with image size $224 \times 224$. GEMax takes an average of 0.22s per image when $\tau=20$, and increases about 0.016s for every interval $\Delta\tau=10$. As a comparison, the time of using full depth range search without $\tau$ (or $\tau=N$) is an unacceptable 33.9s per image with very little entropy gains (7.993 vs. 7.987), which validates the high efficiency of introducing the depth distortion constraint $\tau$.

\subsection{Limitation}
\label{sec:4.6}
Although satisfying results are achieved, semantic information is not considered in the proposed map generation. This may cause the negative enhancement of some facial parts that give more information about the entire face, but contribute less to the discriminative expression feature. For example, Fear is often misclassified as Disgust with a False Negative (FN) rate of $14.31\%$ because chasing the maximal entropy of the full image (eyes + mouth) not only enhancing the discriminative regions (eyes) but also augmenting the relatively confusing parts (mouth).

\section{Ablation Study}
\label{sec:5}
\textbf{Depth range section.}
The effectiveness of the diverse depth range from $d_{min}=5$ cm to $d_{max} =14$ are evaluated in Tab. \ref{tab:Ab_DepthSelection}. Here, 5 cm is the minimum depth to maintain the basic face structure and 14 cm is the maximum depth provided by the dataset. Data with larger $d_i$ have more dynamic range.

\begin{table}[htbp]
\begin{spacing}{1.3}
\small
\begin{center}
\setlength{\tabcolsep}{1.45mm}{
    \caption{Ablation study of depth range selection $d_i$.}
\label{tab:Ab_DepthSelection}
\begin{tabular}{cccccc}
    \toprule[1.5pt]
    Dataset & \multicolumn{5}{c}{Depth Range $d_{i}$ /  Acc. }\\
    \hline
    \multirow{4}{1.5cm}{\centering BU-3DFE}
    & 5 cm  & 6 cm  & 7 cm  & 8 cm  & 9 cm \\
    \cline{2-6}
    & 84.53\% & \bf 84.92\% & 84.37\% & 83.79\% & 83.64\%\\
    \cline{2-6}
    & 10 cm & 11 cm & 12 cm & 13 cm & 14 cm\\
    \cline{2-6}
    & 83.13\% & 82.77\% & 82.67\% & 82.62\% & 82.40\%\\
    \hline

    \multirow{4}{1.5cm}{\centering Bosphorus}
    & 5 cm  & 6 cm  & 7 cm  & 8 cm  & 9 cm \\
    \cline{2-6}
    & 79.31\% & \bf 79.47\% & 79.13\% & 78.92\% & 78.64\%\\
    \cline{2-6}
    & 10 cm & 11 cm & 12 cm & 13 cm & 14 cm\\
    \cline{2-6}
    & 78.64\% & 78.39\% & 78.22\% & 78.17\% & 78.10\%\\
    \bottomrule[1.5pt]
\end{tabular}}
\end{center}
\end{spacing}
\vspace{-4mm}
\end{table}

In the table, best results are achieved at $d_i=6$ cm for both datasets, indicating that facial structure within 6 cm depth contains the most distinctive information for expressions. Depth range larger than 6 cm will introduce redundant and expression-irrelevant information, and gradually decreases the recognition rate.

\textbf{Depth distortion constraint.}
Raw point clouds are first uniformly quantified from 32-bit to 16-bit to avoid the infeasible $2^{32}$ bins histogram calculating. Then, we evaluate the model under different constraints with $\tau$ ranging from 16 to 60.

\begin{table}[htbp]
\begin{spacing}{1.3}
\small
\begin{center}
\setlength{\tabcolsep}{0.5mm}{
    \caption{Ablations of depth distortion constraint $\tau$.}
\label{tab:Ab_GEMC}
\begin{tabular}{ccccccc}
    \toprule[1.5pt]
    Dataset & \multicolumn{6}{c}{BU-3DFE} \\
    \hline
    $\tau$ & 16 & 20  & 30  & 40  & 50  & 60 \\
    Acc. & 84.92\% & \bf 85.81\% & 85.74\% & 85.33\% & 85.26\% & 85.25\%\\
    Entropy & 7.61 & 7.81 & 7.93 & 7.96 & 7.98 & \bf 7.99 \\
    \hline
    Dataset & \multicolumn{6}{c}{Bosphorus} \\
    \hline
    $\tau$ & 16 & 20  & 30  & 40  & 50  & 60 \\
    Acc. &79.47\% & \bf 80.12\% & 80.09\% & 79.83\% & 79.64\% & 79.64\%\\
    Entropy & 7.55 & 7.77 & 7.91 & 7.95 & 7.97 & \bf 7.98 \\
    \bottomrule[1.5pt]
\end{tabular}}
\end{center}
\end{spacing}
\vspace{-4mm}
\end{table}

\begin{figure}[htbp]
\begin{center}
    \includegraphics[width=\linewidth]{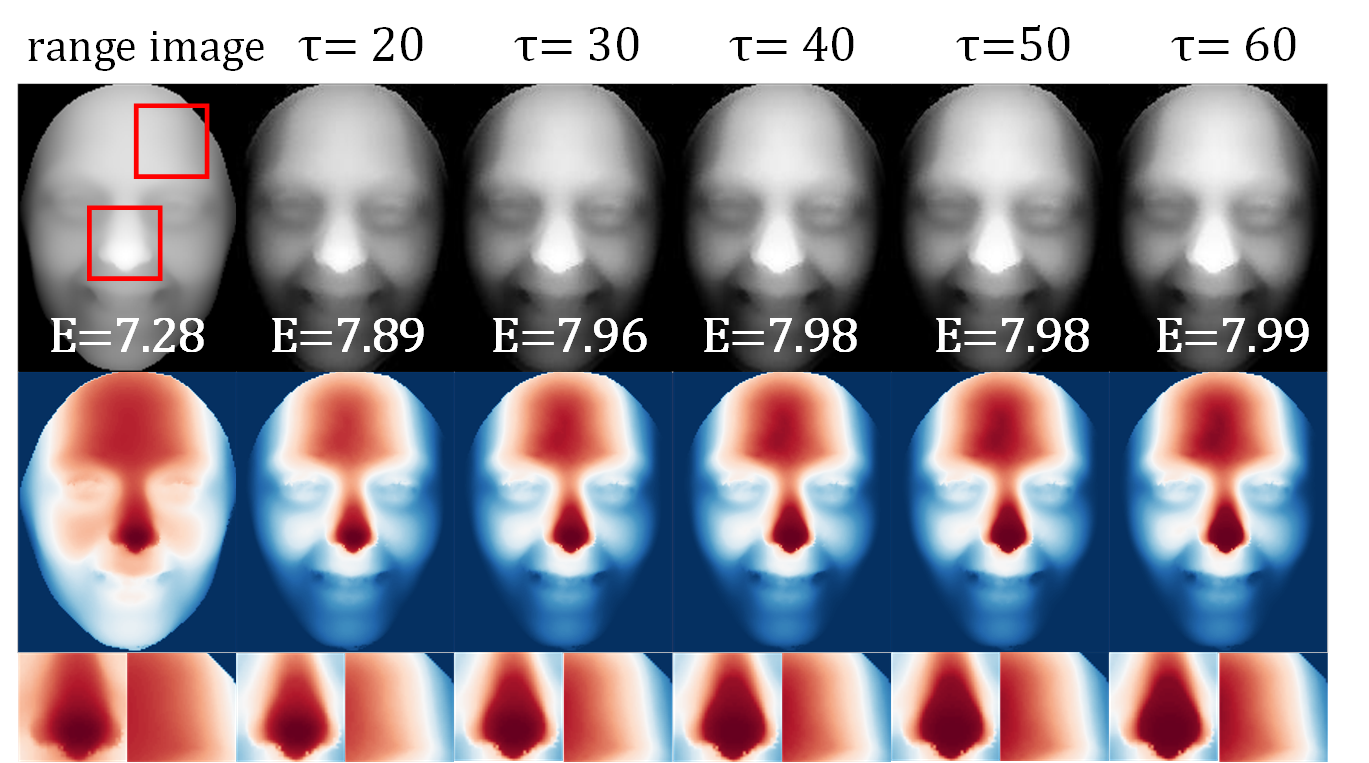}
\end{center}
   \caption{Ablation study of depth distortion constraint $\tau$. Left column: range image. Right columns: GEMax with different $\tau$. Larger $\tau$ achieves higher entropy (E) with loose constraint, but may cause the over enhancement of local facial parts.}
\label{fig:Ab_tau}
\end{figure}

In Tab. \ref{tab:Ab_GEMC}, lower $\tau$ value means stronger constraint. Hence, $\tau=16$ indicates the uniform mapping from $N=2^{16}=4096$ to $K=256$ bins with minimal entropy. Larger $\tau$ relaxes the constraint and produces more entropy gain, but it may also over-enhance the image by chasing the maximal output entropy as shown in Fig. \ref{fig:Ab_tau}, where the contrast of local regions such as the nose and the forehead are often over promoted, leading to geometric deformations.

\begin{table}[htbp]
\begin{spacing}{1.3}
\begin{center}
    \caption{Evaluation of multi-scale learning. Facial parts generated from the proposed FA module are discriminative, and multi-scale learning with local-global images achieves the best performances.}
\label{tab:Ab_FAmodules}
\begin{tabular}{p{1.3cm}<{\centering}p{1.8cm}<{\centering}p{1.0cm}<{\centering}p{1.0cm}<{\centering}p{1.0cm}<{\centering}}
    \toprule[1.5pt]
    Dataset & Input & Texture & Depth & Both \\
    \hline
    \multirow{2}{1.5cm}{\centering BU-3DFE}
    & Full image & 86.79\% & 86.87\% & \multirow{2}{2cm}{ 89.72\%} \\
    & Facial part & 76.39\% & 75.43\% &\\
    \hline
    \multirow{2}{1.5cm}{\centering Bosphorus}
    & Full image & 79.64\% & 81.42\% & \multirow{2}{2cm}{ 83.63\%} \\
    & Facial part & 68.17\% & 73.21\% &\\
    \bottomrule[1.5pt]
\end{tabular}
\end{center}
\end{spacing}
\vspace{-8mm}
\end{table}

\textbf{Multi scale learning.} Multi-scale inputs including full images and facial parts from the FA module are tested in Tab. \ref{tab:Ab_FAmodules}. Facial parts from the FA module only describe a local neighbor structure of the entire face, but recognizing rate from facial parts without complete face information can still achieve a satisfying accuracy (within 10\% drop), demonstrating the effectiveness of FA module. By jointly considering both full images and facial parts, the overall performance using  multi-scale learning can be further boosted.

\section{Conclusion}
In this paper, we have proposed a novel map generation method and a Facial Attention structure for 2D+3D FER task. Taking full advantage of the two improvements, slight expression variations from the discriminative dynamic range are comprehensively enhanced by maximizing the entropy under the depth distortion constraint, and local facial parts can be efficiently generated for multi-scale feature learning. Extensive experiments on different datasets produce state-of-the-art results, demonstrating the generalization ability and the effectiveness of our proposed model. In the future, we will extend the approach to enhance more challenging HDR data, such as infra-red or SAR images, and generalize FA module to 4D (3D + video) FER by exploring the temporal face structure information.

{\small
\bibliographystyle{ieee_fullname}
\bibliography{mainFER}
}

\end{document}